\pdfoutput=1

\documentclass[11pt]{article}

\usepackage[final]{acl}
\usepackage{diagbox}
\usepackage{multirow}
\usepackage{times}
\usepackage{tabularx}
\usepackage{arydshln}
\usepackage{latexsym}
\usepackage{amsmath}
\usepackage{graphicx}
\usepackage{array}
\usepackage{multirow}
\usepackage{CJK}
\usepackage{listings}
\usepackage[T1]{fontenc}

\usepackage[utf8]{inputenc}

\usepackage{microtype}

\usepackage{inconsolata}

\usepackage{graphicx}

\definecolor{color1}{rgb}{0.1,0.7,0.8}
\definecolor{color2}{rgb}{0.9,0.1,0.1}
\definecolor{color3}{rgb}{0.7,0.3,0.7}
\definecolor{color4}{rgb}{0.3,0.3,0.7}
\definecolor{color5}{RGB}{8, 102, 3}
\definecolor{color6}{rgb}{0.53, 0.66, 0.42}

\title{Evaluating Knowledge-based Cross-lingual Inconsistency \\ in Large Language Models}

\author{%
    Xiaolin Xing$^1$\quad Zhiwei He$^2$\quad Haoyu Xu$^1$\quad Xing Wang$^3$$^*$\quad Rui Wang$^2$\quad Yu Hong$^1$\thanks{Yu Hong and Xing Wang are co\-corresponding authors.}\\
    $^1$Soochow University, Computer Science and Technology, Suzhou, China \\
    $^2$Shanghai Jiao Tong University\ \ \ $^3$Tencent AI Lab\\
    \texttt{\small{\{actuallyxxl,tianxianer\}}@gmail.com} \ \ \ \texttt{\small{brightxwang@tencent.com}} 
}

\begin{document}
\begin{CJK}{UTF8}{gkai}
\maketitle
\begin{abstract}
This paper investigates the cross-lingual inconsistencies observed in Large Language Models (LLMs), such as ChatGPT, Llama, and Baichuan, which have shown exceptional performance in various Natural Language Processing (NLP) tasks. Despite their successes, these models often exhibit significant inconsistencies when processing the same concepts across different languages. This study focuses on three primary questions: the existence of cross-lingual inconsistencies in LLMs, the specific aspects in which these inconsistencies manifest, and the correlation between cross-lingual consistency and multilingual capabilities of LLMs.To address these questions, we propose an innovative evaluation method for Cross-lingual Semantic Consistency (xSC) using the LaBSE model. We further introduce metrics for Cross-lingual Accuracy Consistency (xAC) and Cross-lingual Timeliness Consistency (xTC) to comprehensively assess the models' performance regarding semantic, accuracy, and timeliness inconsistencies. By harmonizing these metrics, we provide a holistic measurement of LLMs' cross-lingual consistency. Our findings aim to enhance the understanding and improvement of multilingual capabilities and interpretability in LLMs, contributing to the development of more robust and reliable multilingual language models\footnote{All code and data released at \url{https://github.com/Xingxl2studious/Cross-lingual-Consistency}}.
\end{abstract}

\section{Introduction}

In recent years, the rapid development of Large Language Models (LLMs) has significantly propelled advancements in Natural Language Processing (NLP), exemplified by models such as ChatGPT\footnote{https://chat.openai.com/}, Llama~\cite{touvron2023llama}, and Baichuan~\cite{yang2023baichuan}. These models have demonstrated exceptional performance across a variety of NLP tasks, including machine translation~\cite{jiao2023chatgpt} and question answering~\cite{bang2023multitask}. However, as LLMs are increasingly applied globally, issues of consistency and accuracy in processing multilingual information have become more pronounced.

\begin{figure}[t]
\centering
\includegraphics[width=8cm]{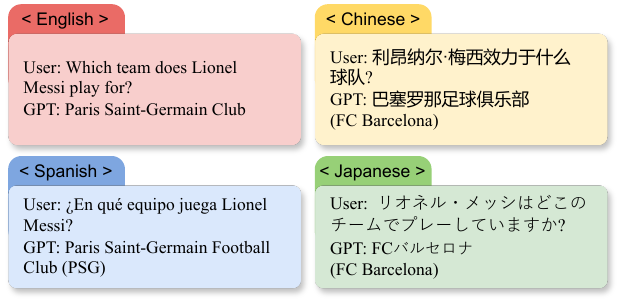}
\caption{Cross-Lingual Inconsistencies in LLM Responses.}
\label{fig:example}
\end{figure}

Multilingual LLMs are designed to break down language barriers, enabling users from different linguistic backgrounds to access high-quality information services. 
Yet, in practice, these models often show notable inconsistencies when dealing with the same concepts across different languages. For instance, as illustrated in Figure~\ref{fig:example}, GPT-3.5-turbo-0325 provided the correct answer, ``Paris Saint-Germain Club (PSG)'' to the question ``Which team does Lionel Messi play for?'' posed in English and Spanish. However, when the same question was asked in Chinese and Japanese, the model incorrectly responded with ``FC Barcelona'' despite Messi's transfer to PSG.

Such cross-lingual inconsistencies are not limited to factual knowledge queries but may also encompass sentiment analysis, named entity recognition, semantic understanding, and other aspects. Consequently, this paper aims to investigate and evaluate the consistency of LLMs in cross-lingual processing. We will explore the following three key questions:

\begin{itemize}
  \item Do LLMs exhibit cross-lingual inconsistency?
  \item In what aspects do LLMs' cross-lingual inconsistencies manifest?
  \item Is there a correlation between the cross-lingual consistency performance of LLMs and their multilingual capabilities?
\end{itemize}

To systematically address these questions, we first introduce an innovative method for evaluating Cross-lingual Semantic Consistency (xSC), based on the cross-lingual semantic vector encoding model LaBSE~\cite{labse}. This approach allows us to verify the existence and stability of cross-lingual inconsistencies within LLMs.

Furthermore, to better measure models' cross-lingual consistency, we expand upon the proposed metric to address three types of inconsistencies manifested by LLMs across different linguistic environments: semantic inconsistency of responses, accuracy inconsistency of responses, and timeliness inconsistency of responses. We introduce the Cross-lingual Accuracy Consistency metric (xAC) and the Cross-lingual Timeliness Consistency metric (xTC) to more comprehensively assess LLMs' cross-lingual performance regarding knowledge accuracy and timeliness. We harmonize the scores of these three metrics to holistically measure LLMs' cross-lingual consistency capabilities.

Finally, we will explore the relationship between LLMs' cross-lingual inconsistency issues and their multilingual abilities, offering a new perspective for understanding and improving the multilingual capabilities and interpretability of LLMs.

\section{Related Work}

\subsection{Factual Knowledge Probing}

Factual Knowledge Probing
In the field of Natural Language Processing (NLP), Pretrained Language Models (PLMs) have been proven to store a vast array of factual knowledge. \citet{petroni-etal-2019-language} examine the capacity of PLMs to store relational knowledge without fine-tuning. It was found that the BERT model learns some types of facts better than others, indicating the potential of language models as unsupervised open-domain question answering systems. 

\citet{heinzerling-inui-2021-language} explore the feasibility of using Pretrained Language Models (LMs) as Knowledge Bases (KBs). It outlines two critical requirements: the ability to store extensive facts involving numerous entities, and the capability to query these facts using natural language paraphrases. The authors compared three different entity representation methods and demonstrated through experiments that LMs can scale to handle millions of entities and memorize and retrieve a vast amount of facts.

\citet{mittal-etal-2023-mokb6} introduce the first multilingual open knowledge base completion dataset, containing facts from Wikipedia in six languages, including English. The research indicates that integrating information across multiple languages and the translation of facts significantly enhances model performance. However, challenges arise for Multilingual Knowledge Graph Embedding (KGE) models when memorizing facts across languages with different scripts.

\subsection{Knowledge-based Cross-lingual Consistency}

Multilingual consistency is a crucial metric for evaluating the performance consistency of multilingual pretrained language models in predicting factual knowledge across different languages. Recent studies have revealed significant inconsistencies even among large multilingual models across various languages.

\citet{fierro-sogaard-2022-factual} discover that multilingual models, such as mBERT and XLM-R, exhibit inconsistencies in English comparable to monolingual English BERT, but show higher inconsistencies across 45 other languages. This reveals the challenges faced by multilingual PLMs in predicting factual knowledge across languages and underscores the importance of addressing cross-lingual consistency issues when building reliable cross-language knowledge bases.

\citet{qi-etal-2023-cross} introduce a ranking-based consistency metric (RankC) to evaluate cross-lingual knowledge consistency independently of accuracy. The findings suggest that while increasing the model size improves factual probing accuracy in most languages, it does not enhance cross-lingual consistency. Furthermore, when new factual associations are inserted into PLMs through model editing, the new knowledge is only transferred to languages with high English RankC scores.

\section{Cross-lingual Inconsistency in Large Language Models}

In an effort to delve into and address the consistency issues exhibited by large language models (LLMs) when processing multilingual requests, this study has constructed a multilingual aligned knowledge-based question-answering dataset. Building upon this, we introduce a Cross-lingual Semantic Consistency metric (xSC), designed to quantify the inconsistency in knowledge representation across multiple languages in question-answering scenarios.

\subsection{MAKQA dataset}

Acknowledging the limitations of existing datasets such as mOKB6~\cite{mittal-etal-2023-mokb6}, MPARARE~\cite{fierro-sogaard-2022-factual}, and BMLAMA~\cite{qi-etal-2023-cross}, which suffer from a narrow domain focus, an over-reliance on machine translation for expanding language coverage, and data structured in triplets not suitable for LLM inference, we build a Multilingual Aligned Knowledge-based Question-Answering dataset (MAKQA) that includes 12 languages: English (En), German (De), Dutch (Nl), French (Fr), Spanish (Es), Italian (It), Portuguese (Pt), Greek (El), Russian (Ru), Chinese (Zh), Japanese (Ja), and Korean (Ko). This dataset encompasses six major knowledge domains including sports, movies, science, history, geography, and literature.

We utilize Wikidata as the primary data source to establish our dataset. Entity names in English are collected from diverse sources, and through Wikipedia, knowledge triplets associated with these entities are acquired. From these triplets, only those containing key relations are selectively retained. 
We capitalize on the feature that every entity in Wikipedia is logged with its multilingual names, thereby expanding English knowledge triples to multilingual aligned knowledge triples. Notably, we only employ translation engines as supplements for specific language names missing from some entities in Wikipedia when necessary. 
Finally, knowledge triples are transformed into knowledge question-answer pairs using GPT-4~\cite{openai2023gpt4}, resulting in our Knowledge QA dataset.

Detailed statistical information about the dataset is available in Table \ref{tab:data_satistic}, and examples of the dataset are presented in Table \ref{tab:data_example}.

\begin{table}
\centering
\begin{tabular}{lccc}
\hline
\textbf{Domain} & \#Entity  & \#Rel & \#QA pairs \\
\hline
Sports & 50 & 9 & 253 \\
Movie & 49 & 17 & 432 \\
Science & 49 & 12 & 492 \\
History & 45 & 12 & 389 \\
Geography & 94 & 6 & 286 \\
Literature & 50 & 5 & 165 \\
\hdashline 
Timeliness & 129 & 2 & 136 \\
\hline
\end{tabular}
\caption{Satistics of the MAKQA dataset.}
\label{tab:data_satistic}
\end{table}

\begin{table*}
\centering

\renewcommand
\arraystretch{1.3}
\begin{tabular}{lcc}
\hline
\textbf{Language}  & \textbf{Question} & \textbf{Answer} \\
\hline
English (En) & In which country is Buenos Aires located? & Argentina \\
Chinese (Zh) & 布宜诺斯艾利斯属于哪个国家？ & 阿根廷\\
German (De) & In welchem Staat liegt Buenos Aires? & Argentinien\\
Dutch (Nl)& In welk land ligt Buenos Aires? & Argentinië\\
Japanese (Ja) & ブエノスアイレスはどの国にありますか？ & アルゼンチン\\
\hline
\end{tabular}
\caption{MAKQA geographical domain showcase.}
\label{tab:data_example}
\end{table*}

\subsection{Cross-lingual Semantic Consistency metric}

The Cross-lingual Semantic Consistency (xSC) evaluation method is designed to assess the degree of knowledge consistency across different languages in Large Language Models (LLMs). Specifically, this metric examines whether a model can provide semantically consistent responses to the same question posed in different languages, thereby evaluating the uniformity of knowledge storage and expression within LLMs across various languages.

To measure this, the method employs the multilingual semantic encoding model LASER to encode the answers generated by LLMs in different languages. It then calculates the cosine similarity distance between these semantic vectors to quantify the model's performance on cross-lingual semantic consistency. The calculation of xSC, as shown in Equation \ref{eq:clsc}, involves prompting the LLM to generate answers in multiple languages, followed by semantic encoding of these answers. It computes the cosine similarity between pairs of languages and averages the similarity across all language combinations to derive the model's xSC score. A score closer to 1 indicates better performance of the model in terms of cross-lingual semantic consistency.

\begin{equation}
\label{eq:clsc}
\begin{aligned}
\text{xSC} &=  \frac{1}{L(L-1)}  \sum_{i=1}^{L}   \sum_{\substack{j=1 \\ j \neq i}}^{L} \text{C}_{i,j}   \\
\text{C}_{i,j} &= \frac{1}{N}  \sum_{s=1}^{N} \text{Cos}(\text{emb}^i_s, \text{emb}^j_s)  \\
\text{emb}^i_s &= \text{LaBSE}(\text{ans}^i_s) 
\end{aligned}
\end{equation}

In the formula, \(\text{ans}_s^i\) represents the answer given by the LLM to the \(s\)th question in the \(i\)th language. \(L\) and \(N\) denote the total number of languages and the total number of QA pairs in the dataset.

\subsection{Experiments}

To comprehensively evaluate the performance of LLMs in cross-lingual knowledge consistency, this study tested five representative LLMs, including the closed-source model GPT-3.5 and four open-source models: Bloomz~\cite{bloomz}, Llama2~\cite{llama2}, Baichuan2~\cite{baichuan2}, and Mistral~\cite{mistral, jiang2024mixtral}. In addition, to determine the upper limit of model performance, we also calculated the xSC score for the actual answers (Groundtruth), which serves as a reference for the ideal state, denoted as Oracle.

In the experiments, we used the LLaMA-Factory framework\footnote{https://github.com/hiyouga/LLaMA-Factory} to build the LLM's API call interface, replicating the LLM's performance in real-world application scenarios. To minimize the impact of the model's ability to follow instructions, we employed a 5-shot context learning strategy, providing five relevant examples prior to inference to aid the LLM in better understanding the task requirements. For each domain, the experiment randomly selected five reference examples from 20 curated examples. All experiments were conducted on servers equipped with four NVIDIA A100-PCIE-40GB GPUs.

\begin{table}
\centering
\begin{tabular}{lc}
\hline
\textbf{Model} & \textbf{Score} \\
\hline
\textbf{Oracle} & \textbf{0.849} \\
\hline
GPT-3.5 & 0.706 \\
Bloomz-7b &  0.414  \\
Llama2-7b & 0.577 \\
Baichuan2-7b & 0.530 \\
Mistral-7b & 0.527 \\
\hline
\label{tab:clc_main_data}
\end{tabular}
\caption{LLMs' cross-lingual semantic consistency score.}
\end{table}

\subsection{Main Result}

As shown in Table 5-3, various large language models (LLMs) exhibit significant differences in their Cross-lingual Semantic Consistency (xSC) scores. The proprietary model GPT-3.5 leads all open-source models with a score of 0.706, demonstrating its superior capability in handling cross-lingual issues. Among the open-source models, Llama2-7b scores 0.577, outperforming other models of similar size, yet still trailing behind GPT-3.5. It is also noted that both proprietary and open-source models, when compared to an ideal state (i.e., the Oracle), have a considerable gap. This outcome reveals substantial room for improvement, especially in open-source models, in terms of cross-lingual consistency.

\subsection{Analysis}

Furthermore, to test the stability of cross-lingual inconsistency issues in LLMs, we conduct further experiments from two dimensions: domain differences and prompt design.

\begin{table*}
\centering
\begin{tabular}{lcccccc}
\hline
\multirow{2}{*}{\bf Model} & \multicolumn{6}{c}{\bf Domain }\\
 &\textbf{Sports}  &\textbf{Movie} & \textbf{Science} & \textbf{History}& \textbf{Geography} & \textbf{Literature}  \\
\hline
\textbf{Oracle} & \textbf{0.834} & \textbf{0.870} & \textbf{0.858} & \textbf{0.817} & \textbf{0.866} & \textbf{0.838} \\
\hline
GPT-3.5	 & 0.767 & 0.647 & 0.691 & 0.678 & 0.804 & 0.721 \\
Bloomz-7b	 & 0.455 & 0.332 & 0.412 & 0.390 & 0.558 & 0.379  \\
Llama2-7b	  & 0.579 & 0.511 & 0.657 & 0.528 & 0.661 & 0.476 \\
Baichuan2-7b   & 0.588 & 0.427 & 0.536 & 0.519 & 0.653 & 0.511 \\
Mistral-7b	  & 0.561 & 0.484 & 0.559 & 0.521 & 0.566 & 0.438 \\
\hline
\end{tabular}
\caption{Cross-lingual semantic consistency score in different domains.}
\label{tab:domain_satistic}
\end{table*}

\paragraph{Domain-Specific Analysis} In this experiment, we independently evaluate the performance of five representative models across six different domains using xSC, as detailed in Table \ref{tab:domain_satistic}. The results indicate that despite fluctuations in scores across various domains, these fluctuations do not significantly affect the overall trend of cross-lingual semantic consistency. GPT-3.5 consistently shows a leading advantage in all domains, while Bloomz-7b generally lags behind other models in each domain. Among the open-source models, Llama2-7b performs best in four out of six domains. These findings suggest that while there are significant knowledge differences between domains, such differences do not materially affect the xSC scores of LLMs. In other words, a model that performs well maintains high cross-lingual consistency across different domains, indicating that the issue of cross-lingual inconsistency is an inherent and stable behavior of the model, independent of specific knowledge domains.

\begin{table}
\centering
\begin{tabular}{lccc}
\hline
\textbf{Models} & \textbf{Prompt1} & \textbf{Prompt2} & \textbf{Prompt3} \\
\hline
Bloomz-7b & 0.414 & 0.417 & 0.426 \\
Llama2-7b & 0.577 & 0.552 & 0.562 \\
Baichuan2-7b & 0.530 & 0.534 & 0.519 \\
Mistral-7b & 0.527 & 0.523 & 0.518\\
\hline
\end{tabular}
\caption{Cross-lingual semantic consistency score with different prompts.}
\label{tab:prompt-sensitive}
\end{table}

\paragraph{Prompt Design Analysis} This experiment compares whether LLMs exhibit significant fluctuations in cross-lingual consistency when facing the same question posed by different prompts. In addition to the original question (Prompt 1), we construct two new sets of prompts for the experiment. Specifically, Prompt 2 employs a standardized question template, generating standard questions by filling in key entities and relations; Prompt 3 derives from GPT-4's adaptation of the original question. Table \ref{tab:prompt-sensitive} shows the performance of five representative models under these different prompts. Although there are subtle differences in model performance based on different prompts, such as Bloomz-7b scoring 0.414, 0.417, and 0.426 under the three prompts, these variations do not alter the overall ranking and score differences between models. This further confirms that the issue of cross-lingual consistency in LLMs is a stable model behavior, not affected by different prompt designs, and also validates the robustness of the xSC metric.

\section{Manifestations of Cross-lingual Inconsistency}

In the previous sections, we demonstrated through the Cross-Lingual Semantic Consistency (xSC) metric that Large Language Models (LLMs) exhibit significant cross-lingual semantic inconsistencies when handling requests in different languages. However, semantic inconsistency is just one form of cross-lingual inconsistency. As shown in Figure \ref{fig:example}, the responses of the model in various languages not only differ semantically but also show discrepancies in accuracy consistency (i.e., whether the model provides the same correct or incorrect answer across languages) and timeliness consistency (i.e., whether the model provides timely answers across different languages). Therefore, to more comprehensively evaluate the cross-lingual consistency performance of the model, we further propose the Cross-Lingual Accuracy Consistency metric (xAC) and Cross-Lingual Timeliness Consistency metric (xTC). These are then combined with xSC to obtain the overall Cross-Lingual Consistency metric (xC).

\begin{table*}
\centering
\begin{tabular}{l r cccc}
\hline
\multirow{2}{*}{\bf Model} & \multirow{2}{*}{\bf 
 Size} &  \multicolumn{4}{c}{\bf Metric} \\
& & \textbf{xSC} & \textbf{xAC} & \textbf{xTC} & \bf xC \\
\hline
\textsc{GPT-3.5} & -- & \textbf{0.706} & \textbf{0.489} & \textbf{0.508} & \textbf{0.552}\\

\hline
\multirow{4}{*}{ \textsc{Bloomz}} & 0.6B &  0.353 & 0.261 & 0.236 & 0.275\\
 & 1B & 0.389 & 0.256 & 0.199 & 0.260\\
 & 3B & 0.409 & 0.298 & 0.191 & 0.272\\
 & 7B &  0.414 & 0.275 & 0.193 & 0.267\\

\hdashline 
\multirow{2}{*}{ \textsc{Llama2}} & 7B & 0.577 & 0.243 & 0.297 & 0.326\\
  & 13B & 0.563 & 0.293 & 0.321 & 0.361\\

\hdashline
\multirow{2}{*}{ \textsc{Baichuan2}} & 7B &0.530 & 0.342 & 0.413 & 0.415\\
  & 13B &  0.564  & 0.367 &  0.391 & 0.425\\

\hdashline
\textsc{Mistral} & 7B & 0.527 & 0.245 & 0.349 & 0.339\\
\textsc{Mixtral} & 8x7B & 0.666 & 0.430 & 0.450 & 0.496\\

\hline
\end{tabular}
\caption{\label{main_data}
The main result of assessing the cross-lingual consistency of LLMs.
}
\end{table*}

\subsection{Cross-lingual Accuracy Consistency metric}

The Cross-lingual Accuracy Consistency (xAC) metric aims to assess whether the answers provided by LLMs to multilingual knowledge queries are consistently accurate. Cross-lingual accuracy reflects the model's ability to perform downstream tasks in different language environments and is directly related to its multilingual generalization capability, making it a core metric for evaluating multilingual performance. By evaluating the consistency of cross-lingual accuracy, this method reveals whether the model can handle multilingual queries with stable accuracy across language boundaries, which is crucial for assessing the performance of LLMs in multilingual tasks.

We measure the accuracy of responses by calculating the CHRF score~\cite{chrf++} between the model's answers and the ground truth in each language. Then, we evaluate the correlation between accuracy scores for different language pairs by calculating the Spearman rank correlation coefficient for all accuracy scores across languages. The average correlation score across all language pairs serves as the metric for cross-lingual accuracy consistency, calculated as follows:

\begin{equation}
\label{eq:clac}
\begin{aligned}
\text{xAC} &=  \frac{1}{L(L-1)}  \sum_{i=1}^{L}   \sum_{\substack{j=1 \\ j \neq i}}^{L} \text{C}^A_{i,j}    \\
\text{C}^A_{i,j} &= Spearman(\text{acc}^i, \text{acc}^j) \\
\text{acc}^i_t &= \text{CHRF}(\text{ans}^i_t, y_t),  \\
&\ \ \ \ \ \ \ \ \ \ \ \ \ \ \ for\ t = 1,2,...,n
\end{aligned}
\end{equation}

\subsection{Cross-lingual Timeliness Consistency metric}

The Cross-lingual Timeliness Consistency (xTC) metric aims to evaluate the consistency of LLMs in answering multilingual knowledge queries that are sensitive to timeliness. Ideally, LLMs should provide synchronously updated information for the same time-sensitive query posed in different languages. As shown in Figure \ref{fig:example}, when querying recent news events or knowledge, the responses of LLMs differ in timeliness across languages. The xTC metric not only assesses the model's cross-lingual timeliness consistency in time-critical scenarios but also helps in analyzing the model's internal knowledge consistency regarding timeliness across languages.

The xTC evaluation method focuses on the model's performance in handling time-sensitive queries. Since regular queries do not involve timeliness changes, we use a specially designed dataset of time-sensitive questions, with statistical information shown in Table \ref{tab:data_satistic}. This dataset consists of a series of highly time-sensitive questions, each with multiple candidate answers ranked by timeliness to test the model's ability to grasp the latest information. The evaluation process is similar to xAC and includes the following four steps:

First, we calculate the CHRF score between the model's answer and a set of candidate answers with different timeliness to determine the best matching candidate answer and its timeliness ranking $r$. Next, based on the ranking $r$, we calculate a timeliness score for each answer, defined as the reciprocal of the timeliness ranking $1/r$ multiplied by the CHRF score, to quantify the timeliness of the model's answer for a specific question. The higher the score (closer to 1), the more up-to-date the model's answer is; the lower the score, the more outdated the answer is. If the model fails to provide a correct answer, the score is zero. Subsequently, we calculate the Spearman rank correlation coefficient for the timeliness scores across different language pairs to assess the model's cross-lingual timeliness consistency. 
Finally, by averaging the Spearman correlation coefficients across all language pairs, we obtain the model's overall xTC score:

\begin{equation}
\label{eq:cltc}
\begin{aligned}
\text{xTC} &=  \frac{1}{L(L-1)}  \sum_{i=1}^{L}   \sum_{\substack{j=1 \\ j \neq i}}^{L} \text{C}^A_{i,j}    \\
\text{C}^A_{i,j} &= Spearman(\text{Tscore}^i, \text{Tscore}^j) \\
\text{Tscore}^i_t &=\frac{ \text{max}_{r}\  \text{CHRF}(\text{ans}^i_t, y_{t, r})}{R},  \\
&\ \ \ \ \ \ \ \ \ \ \ \ \ \ \ \ \ \ \ \ \ \ \ \ \ \ \ for\ t = 1,2,...,n
\end{aligned}
\end{equation}

In the formula, $\text{Tscore}^i_t$ denotes the timeliness score of answer $t$ in language $i$. $R$ signifies the maximum possible ranking.

\subsection{Cross-lingual Consistency metric }

After obtaining the xSC, xAC, and xTC scores of the LLMs, we compute the harmonic mean of these three scores to derive the model's overall cross-lingual consistency score (xC), thereby comprehensively measuring the cross-lingual consistency performance of the LLMs. The calculation process is as follows:

\begin{equation}
\label{eq:clc}
\text{xC} = \frac{3}{\frac{1}{\text{xSC}} + \frac{1}{\text{xAC}} + \frac{1}{\text{xTC}}}
\end{equation}

\subsection{Experiments}

We adopt the same experimental setup as previously described. To better illustrate the cross-lingual performance of each model type and to explore the impact of model parameters on cross-lingual performance, we test all versions of each model type with parameters up to 13B.

\subsection{Result}

The experimental results are shown in Table \ref{main_data}. It is evident that different models exhibit significant differences in cross-lingual consistency, with GPT-3.5 performing the best across all metrics. Among the open-source models, Baichuan2 demonstrates good cross-lingual consistency, showing strong performance on all three metrics compared to models of similar size. However, Bloomz lags behind other models in all aspects. Despite using a large multilingual dataset for both pre-training and fine-tuning, this indicates that merely increasing the proportion of multilingual training data does not break the knowledge barriers between languages.

Overall, the performance differences between models are most balanced in semantic consistency (xSC), while accuracy and timeliness consistency (xAC and xTC) are more influenced by external factors, posing higher demands on the models and resulting in more significant differences. Only Mixtral approaches the performance level of GPT-3.5.

Within different models, performance generally improves with an increase in parameters, but the degree and effect of this improvement vary by model. For instance, in the case of Bloomz, the performance gains from increasing parameters (from 0.6B to 7B) are not significant, especially in the xAC and xTC metrics. This suggests that the structure and training data of the Bloomz model have design limitations that cannot be significantly improved by simply increasing the number of parameters. In contrast, Mixtral enhances model parameters using the MOE structure, leading to significant performance improvements across all metrics. In summary, larger datasets and more complex model architectures (such as GPT-3.5 and Mixtral) are effective methods for enhancing cross-lingual consistency.
\section{Relation Between Cross-Lingual Consistency and Translation Capabilities}

This section aims to explore the proposed third question: Is there a correlation between the cross-lingual consistency performance of LLMs and their multilingual capabilities?

We investigate the potential correlation between cross-lingual consistency and multilingual capabilities of LLMs through multilingual translation tasks. Using the Flores-200 development test (devtest) dataset~\cite{flores101, nllb2022}, we selected 12 test languages, creating a comprehensive test set with 132 translation directions. Based on this test set, we evaluated the translation capabilities of two LLMs: Bloomz-7b and Baichuan2-7b. To mitigate the impact of tokenization on translation metrics for certain languages (such as Chinese, Japanese, and Korean), we used the CHRF metric~\cite{chrf++} to quantify the performance of the models in each translation direction.

\begin{figure}[t]
\centering
\includegraphics[width=8cm]{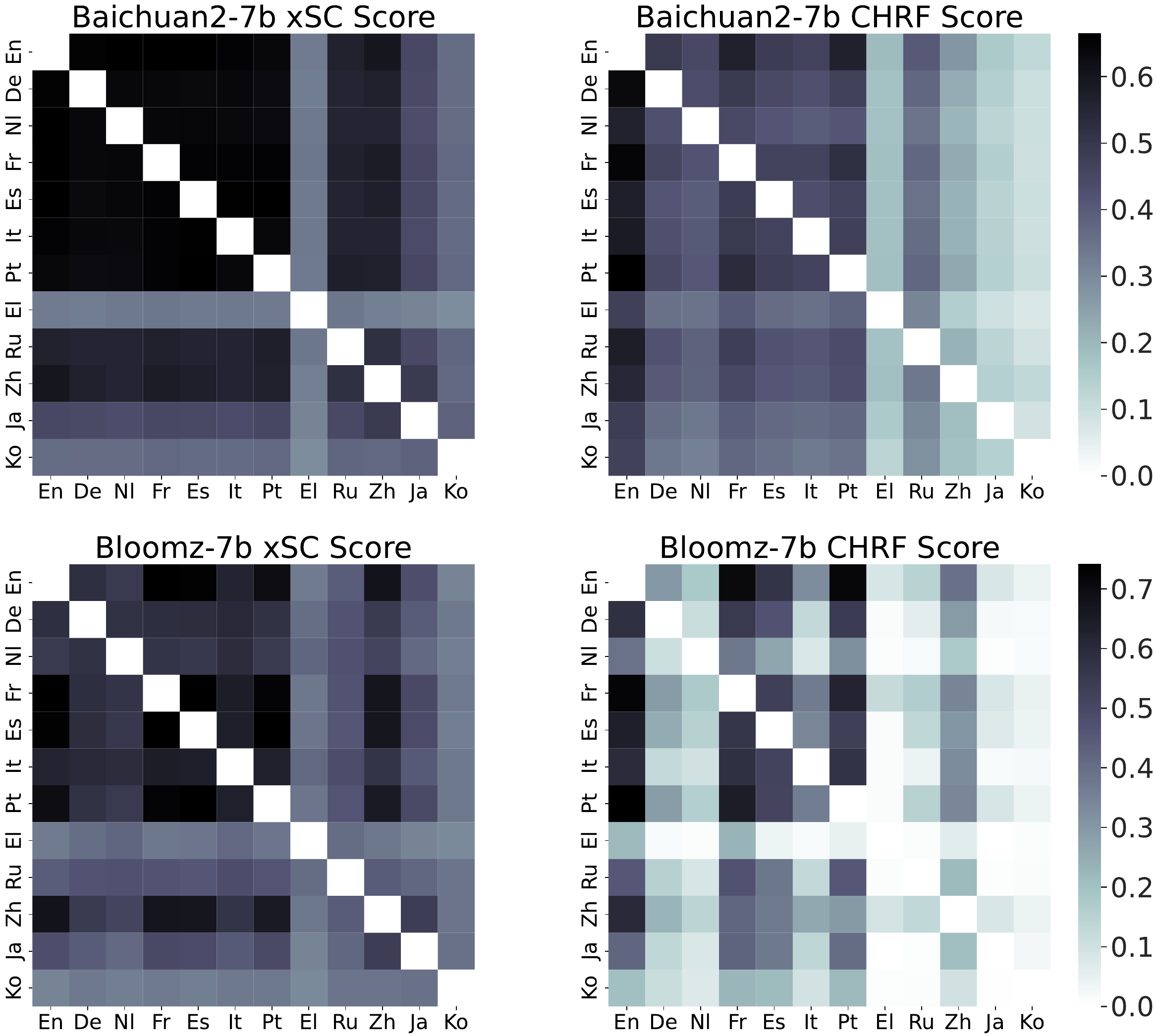}
\caption{LLM performance in multilingual translation and average xSC score distribution.}
\label{fig:translation_heatmap}
\end{figure}

\begin{figure}[t]
\centering
\includegraphics[width=8cm]{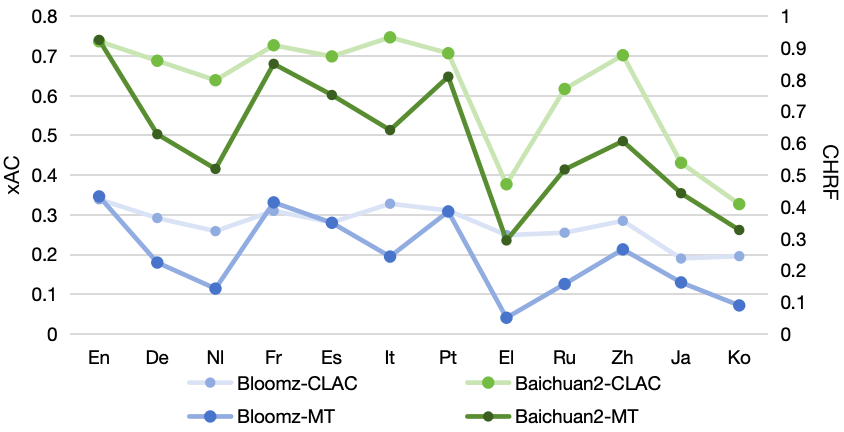}
\caption{LLM performance in multi-language translation and average xAC score distribution.}
\label{fig:mt-xAC}
\end{figure}

\paragraph{Analysis of the Correlation Between Multilingual Translation Performance and Cross-lingual Semantic Consistency (xSC)} The left side of Figure \ref{fig:translation_heatmap} presents two heatmaps showing the distribution of xSC scores between different languages for two models, while the right side displays the zero-shot translation performance scores between different languages. The results indicate a consistent distribution trend between the performance of LLMs in multilingual translation tasks and their xSC scores. Specifically, these models demonstrate higher translation accuracy and cross-lingual semantic consistency in tasks involving Germanic languages (such as English, German, and Dutch) and Indo-Romance languages (such as French, Spanish, Italian, and Portuguese). In contrast, the performance and cross-linguistic consistency are relatively weaker in translation tasks that do not involve these two language families.

\paragraph{Analysis of the Correlation Between Multilingual Translation Performance and Cross-lingual Accuracy Consistency (xAC)} Figure \ref{fig:mt-xAC} explores the correlation between the multilingual translation capabilities of LLMs and xAC. Each data point in the figure represents the model's average performance score for tasks centered on that language. Darker points indicate the model's average performance across all translation tasks involving that particular language, while lighter points correspond to the model's average xAC score for that language. The results show a clear positive correlation between the multilingual translation capabilities of LLMs and their average xAC scores. This correlation is consistent not only across different models, indicating that the higher the average xAC score, the stronger the overall multilingual translation performance, but also within the same model across different languages, showing that the higher the average xAC score for a particular language, the stronger the model's average performance in all translation tasks centered on that language.

The positive correlation observed between the xSC and xAC scores and the translation performance suggests that enhancing cross-lingual consistency could be a viable strategy to improve multilingual capabilities of LLMs. Future research could further explore this correlation by including a more diverse set of languages and examining the underlying factors that contribute to cross-lingual consistency. By continuing to refine and test these models, we can better understand the intricacies of multilingual translation and develop LLMs that are more robust and accurate across a wide range of languages.

\section{Conclusion}

Our research attempts to address the following three key questions:

\paragraph{Do LLMs exhibit cross-lingual inconsistency?}
To verify the presence of cross-lingual inconsistency in models, we construct a Multilingual Aligned Knowledge-based Question-Answering dataset (MAKQA). Using this dataset, we introduce the Cross-lingual Semantic Consistency metric (xSC) and assess five advanced LLMs, demonstrating significant cross-lingual inconsistencies by comparing their scores with those of an ideal state (Oracle). Our experiments consistently confirm the presence of this issue.

\paragraph{In what aspects does cross-lingual inconsistency manifest within LLMs?} 
By analyzing the performance of existing models, we supplement the xSC with the Cross-lingual Accuracy Consistency metric (xAC) and the Cross-lingual Timeliness Consistency metric (xTC). By harmonically averaging these three metrics, we provide a comprehensive assessment of cross-lingual inconsistency in LLMs. Our findings indicate that these inconsistencies manifest not only in semantic understanding but also in accuracy and timeliness, underscoring the multifaceted nature of this issue.

\paragraph{Is there a relationship between the cross-lingual consistency of LLMs and their multilingual capabilities?}
Our experiments validate a positive correlation between the models' cross-lingual consistency and their multilingual translation abilities, grounded in multilingual translation tasks. This suggests that improvements in multilingual translation capabilities can enhance cross-lingual consistency, offering a potential pathway for mitigating the inconsistencies observed.

\section{Limitations}

This study is dedicated to exploring how Large Language Models (LLMs) perform in terms of cross-lingual consistency. We have selected factual knowledge-based question-and-answer tasks as our evaluative instrument and have experimented with five distinct LLMs across a dozen languages. It is important to highlight that while such question-and-answer tasks can benefit from enhanced performance through Retrieval-augmented Generation (RAG), the true test for LLMs lies in scenarios that require reliance on their internal knowledge bases to address indirect queries. Our research, therefore, zeroes in on these types of tasks intending to evaluate and foster the consistency and precision with which LLMs handle cross-lingual information.

However, the MAKQA dataset currently only supports 12 languages, most of which are resource-rich. Given the limited performance of LLMs in low-resource languages, we think that the current collection of languages is sufficient to preliminarily demonstrate the model's cross-lingual consistency among common languages. In the future, we plan to expand the dataset to include more language support, especially for those languages that are less resourced, to more comprehensively evaluate the cross-lingual capabilities of LLMs.

Another limitation of this paper is that our work is confined to assessing and analyzing the issue of cross-lingual consistency in LLMs. In future research, we will strive to explore how to enhance the cross-lingual consistency of LLMs with lower resource consumption. This effort is not only to address the inconsistencies LLMs exhibit when processing different languages but also to provide more stable and reliable support in practical application scenarios. We anticipate that these efforts will aid in building intelligent systems without language boundaries.

\bibliography{custom}

\begin{thebibliography}{19}
\providecommand{\natexlab}[1]{#1}

\bibitem[{Baichuan(2023)}]{baichuan2}
Baichuan. 2023.
\newblock \href {https://arxiv.org/abs/2309.10305} {Baichuan 2: Open large-scale language models}.
\newblock \emph{arXiv preprint arXiv:2309.10305}.

\bibitem[{Bang et~al.(2023)Bang, Cahyawijaya, Lee, Dai, Su, Wilie, Lovenia, Ji, Yu, Chung et~al.}]{bang2023multitask}
Yejin Bang, Samuel Cahyawijaya, Nayeon Lee, Wenliang Dai, Dan Su, Bryan Wilie, Holy Lovenia, Ziwei Ji, Tiezheng Yu, Willy Chung, et~al. 2023.
\newblock A multitask, multilingual, multimodal evaluation of chatgpt on reasoning, hallucination, and interactivity.
\newblock \emph{arXiv preprint arXiv:2302.04023}.

\bibitem[{Feng et~al.(2022)Feng, Yang, Cer, Arivazhagan, and Wang}]{labse}
Fangxiaoyu Feng, Yinfei Yang, Daniel Cer, Naveen Arivazhagan, and Wei Wang. 2022.
\newblock \href {https://doi.org/10.18653/V1/2022.ACL-LONG.62} {Language-agnostic {BERT} sentence embedding}.
\newblock In \emph{Proceedings of the 60th Annual Meeting of the Association for Computational Linguistics (Volume 1: Long Papers), {ACL} 2022, Dublin, Ireland, May 22-27, 2022}, pages 878--891. Association for Computational Linguistics.

\bibitem[{Fierro and S{\o}gaard(2022)}]{fierro-sogaard-2022-factual}
Constanza Fierro and Anders S{\o}gaard. 2022.
\newblock \href {https://doi.org/10.18653/v1/2022.findings-acl.240} {Factual consistency of multilingual pretrained language models}.
\newblock In \emph{Findings of the Association for Computational Linguistics: ACL 2022}, pages 3046--3052, Dublin, Ireland. Association for Computational Linguistics.

\bibitem[{Goyal et~al.(2021)Goyal, Gao, Chaudhary, Chen, Wenzek, Ju, Krishnan, Ranzato, Guzm\'{a}n, and Fan}]{flores101}
Naman Goyal, Cynthia Gao, Vishrav Chaudhary, Peng-Jen Chen, Guillaume Wenzek, Da~Ju, Sanjana Krishnan, Marc'Aurelio Ranzato, Francisco Guzm\'{a}n, and Angela Fan. 2021.
\newblock The flores-101 evaluation benchmark for low-resource and multilingual machine translation.

\bibitem[{Heinzerling and Inui(2021)}]{heinzerling-inui-2021-language}
Benjamin Heinzerling and Kentaro Inui. 2021.
\newblock \href {https://doi.org/10.18653/v1/2021.eacl-main.153} {Language models as knowledge bases: On entity representations, storage capacity, and paraphrased queries}.
\newblock In \emph{Proceedings of the 16th Conference of the European Chapter of the Association for Computational Linguistics: Main Volume}, pages 1772--1791, Online. Association for Computational Linguistics.

\bibitem[{Jiang et~al.(2023)Jiang, Sablayrolles, Mensch, Bamford, Chaplot, de~las Casas, Bressand, Lengyel, Lample, Saulnier, Lavaud, Lachaux, Stock, Scao, Lavril, Wang, Lacroix, and Sayed}]{mistral}
Albert~Q. Jiang, Alexandre Sablayrolles, Arthur Mensch, Chris Bamford, Devendra~Singh Chaplot, Diego de~las Casas, Florian Bressand, Gianna Lengyel, Guillaume Lample, Lucile Saulnier, Lélio~Renard Lavaud, Marie-Anne Lachaux, Pierre Stock, Teven~Le Scao, Thibaut Lavril, Thomas Wang, Timothée Lacroix, and William~El Sayed. 2023.
\newblock \href {https://arxiv.org/abs/2310.06825} {Mistral 7b}.
\newblock \emph{Preprint}, arXiv:2310.06825.

\bibitem[{Jiang et~al.(2024)Jiang, Sablayrolles, Roux, Mensch, Savary, Bamford, Chaplot, de~las Casas, Hanna, Bressand, Lengyel, Bour, Lample, Lavaud, Saulnier, Lachaux, Stock, Subramanian, Yang, Antoniak, Scao, Gervet, Lavril, Wang, Lacroix, and Sayed}]{jiang2024mixtral}
Albert~Q. Jiang, Alexandre Sablayrolles, Antoine Roux, Arthur Mensch, Blanche Savary, Chris Bamford, Devendra~Singh Chaplot, Diego de~las Casas, Emma~Bou Hanna, Florian Bressand, Gianna Lengyel, Guillaume Bour, Guillaume Lample, Lélio~Renard Lavaud, Lucile Saulnier, Marie-Anne Lachaux, Pierre Stock, Sandeep Subramanian, Sophia Yang, Szymon Antoniak, Teven~Le Scao, Théophile Gervet, Thibaut Lavril, Thomas Wang, Timothée Lacroix, and William~El Sayed. 2024.
\newblock \href {https://arxiv.org/abs/2401.04088} {Mixtral of experts}.
\newblock \emph{Preprint}, arXiv:2401.04088.

\bibitem[{Jiao et~al.(2023)Jiao, Wang, Huang, Wang, and Tu}]{jiao2023chatgpt}
Wenxiang Jiao, Wenxuan Wang, JT~Huang, Xing Wang, and ZP~Tu. 2023.
\newblock Is chatgpt a good translator? yes with gpt-4 as the engine.
\newblock \emph{arXiv preprint arXiv:2301.08745}.

\bibitem[{Mittal et~al.(2023)Mittal, Kolluru, Chakrabarti, and {Mausam}}]{mittal-etal-2023-mokb6}
Shubham Mittal, Keshav Kolluru, Soumen Chakrabarti, and {Mausam}. 2023.
\newblock \href {https://doi.org/10.18653/v1/2023.acl-short.19} {m{OKB}6: A multilingual open knowledge base completion benchmark}.
\newblock In \emph{Proceedings of the 61st Annual Meeting of the Association for Computational Linguistics (Volume 2: Short Papers)}, pages 201--214, Toronto, Canada. Association for Computational Linguistics.

\bibitem[{Muennighoff et~al.(2022)Muennighoff, Wang, Sutawika, Roberts, Biderman, Scao, Bari, Shen, Yong, Schoelkopf et~al.}]{bloomz}
Niklas Muennighoff, Thomas Wang, Lintang Sutawika, Adam Roberts, Stella Biderman, Teven~Le Scao, M~Saiful Bari, Sheng Shen, Zheng-Xin Yong, Hailey Schoelkopf, et~al. 2022.
\newblock Crosslingual generalization through multitask finetuning.
\newblock \emph{arXiv preprint arXiv:2211.01786}.

\bibitem[{OpenAI et~al.(2023)OpenAI, :, Achiam, Adler, Agarwal, Ahmad, Akkaya, Aleman, Almeida, Altenschmidt, Altman, Anadkat, Avila, Babuschkin, Balaji, Balcom, Baltescu, Bao, Bavarian, Belgum, Bello, Berdine, Bernadett-Shapiro, Berner, Bogdonoff, Boiko, Boyd, Brakman, Brockman, Brooks, Brundage, Button, Cai, Campbell, Cann, Carey, Carlson, Carmichael, Chan, Chang, Chantzis, Chen, Chen, Chen, Chen, Chen, Chess, Cho, Chu, Chung, Cummings, Currier, Dai, Decareaux, Degry, Deutsch, Deville, Dhar, Dohan, Dowling, Dunning, Ecoffet, Eleti, Eloundou, Farhi, Fedus, Felix, Fishman, Forte, Fulford, Gao, Georges, Gibson, Goel, Gogineni, Goh, Gontijo-Lopes, Gordon, Grafstein, Gray, Greene, Gross, Gu, Guo, Hallacy, Han, Harris, He, Heaton, Heidecke, Hesse, Hickey, Hickey, Hoeschele, Houghton, Hsu, Hu, Hu, Huizinga, Jain, Jain, Jang, Jiang, Jiang, Jin, Jin, Jomoto, Jonn, Jun, Kaftan, Łukasz Kaiser, Kamali, Kanitscheider, Keskar, Khan, Kilpatrick, Kim, Kim, Kim, Kirchner, Kiros, Knight, Kokotajlo, Łukasz Kondraciuk,
  Kondrich, Konstantinidis, Kosic, Krueger, Kuo, Lampe, Lan, Lee, Leike, Leung, Levy, Li, Lim, Lin, Lin, Litwin, Lopez, Lowe, Lue, Makanju, Malfacini, Manning, Markov, Markovski, Martin, Mayer, Mayne, McGrew, McKinney, McLeavey, McMillan, McNeil, Medina, Mehta, Menick, Metz, Mishchenko, Mishkin, Monaco, Morikawa, Mossing, Mu, Murati, Murk, Mély, Nair, Nakano, Nayak, Neelakantan, Ngo, Noh, Ouyang, O'Keefe, Pachocki, Paino, Palermo, Pantuliano, Parascandolo, Parish, Parparita, Passos, Pavlov, Peng, Perelman, de~Avila Belbute~Peres, Petrov, de~Oliveira~Pinto, Michael, Pokorny, Pokrass, Pong, Powell, Power, Power, Proehl, Puri, Radford, Rae, Ramesh, Raymond, Real, Rimbach, Ross, Rotsted, Roussez, Ryder, Saltarelli, Sanders, Santurkar, Sastry, Schmidt, Schnurr, Schulman, Selsam, Sheppard, Sherbakov, Shieh, Shoker, Shyam, Sidor, Sigler, Simens, Sitkin, Slama, Sohl, Sokolowsky, Song, Staudacher, Such, Summers, Sutskever, Tang, Tezak, Thompson, Tillet, Tootoonchian, Tseng, Tuggle, Turley, Tworek, Uribe, Vallone,
  Vijayvergiya, Voss, Wainwright, Wang, Wang, Wang, Ward, Wei, Weinmann, Welihinda, Welinder, Weng, Weng, Wiethoff, Willner, Winter, Wolrich, Wong, Workman, Wu, Wu, Wu, Xiao, Xu, Yoo, Yu, Yuan, Zaremba, Zellers, Zhang, Zhang, Zhao, Zheng, Zhuang, Zhuk, and Zoph}]{openai2023gpt4}
OpenAI, :, Josh Achiam, Steven Adler, Sandhini Agarwal, Lama Ahmad, Ilge Akkaya, Florencia~Leoni Aleman, Diogo Almeida, Janko Altenschmidt, Sam Altman, Shyamal Anadkat, Red Avila, Igor Babuschkin, Suchir Balaji, Valerie Balcom, Paul Baltescu, Haiming Bao, Mo~Bavarian, Jeff Belgum, Irwan Bello, Jake Berdine, Gabriel Bernadett-Shapiro, Christopher Berner, Lenny Bogdonoff, Oleg Boiko, Madelaine Boyd, Anna-Luisa Brakman, Greg Brockman, Tim Brooks, Miles Brundage, Kevin Button, Trevor Cai, Rosie Campbell, Andrew Cann, Brittany Carey, Chelsea Carlson, Rory Carmichael, Brooke Chan, Che Chang, Fotis Chantzis, Derek Chen, Sully Chen, Ruby Chen, Jason Chen, Mark Chen, Ben Chess, Chester Cho, Casey Chu, Hyung~Won Chung, Dave Cummings, Jeremiah Currier, Yunxing Dai, Cory Decareaux, Thomas Degry, Noah Deutsch, Damien Deville, Arka Dhar, David Dohan, Steve Dowling, Sheila Dunning, Adrien Ecoffet, Atty Eleti, Tyna Eloundou, David Farhi, Liam Fedus, Niko Felix, Simón~Posada Fishman, Juston Forte, Isabella Fulford, Leo Gao,
  Elie Georges, Christian Gibson, Vik Goel, Tarun Gogineni, Gabriel Goh, Rapha Gontijo-Lopes, Jonathan Gordon, Morgan Grafstein, Scott Gray, Ryan Greene, Joshua Gross, Shixiang~Shane Gu, Yufei Guo, Chris Hallacy, Jesse Han, Jeff Harris, Yuchen He, Mike Heaton, Johannes Heidecke, Chris Hesse, Alan Hickey, Wade Hickey, Peter Hoeschele, Brandon Houghton, Kenny Hsu, Shengli Hu, Xin Hu, Joost Huizinga, Shantanu Jain, Shawn Jain, Joanne Jang, Angela Jiang, Roger Jiang, Haozhun Jin, Denny Jin, Shino Jomoto, Billie Jonn, Heewoo Jun, Tomer Kaftan, Łukasz Kaiser, Ali Kamali, Ingmar Kanitscheider, Nitish~Shirish Keskar, Tabarak Khan, Logan Kilpatrick, Jong~Wook Kim, Christina Kim, Yongjik Kim, Hendrik Kirchner, Jamie Kiros, Matt Knight, Daniel Kokotajlo, Łukasz Kondraciuk, Andrew Kondrich, Aris Konstantinidis, Kyle Kosic, Gretchen Krueger, Vishal Kuo, Michael Lampe, Ikai Lan, Teddy Lee, Jan Leike, Jade Leung, Daniel Levy, Chak~Ming Li, Rachel Lim, Molly Lin, Stephanie Lin, Mateusz Litwin, Theresa Lopez, Ryan Lowe,
  Patricia Lue, Anna Makanju, Kim Malfacini, Sam Manning, Todor Markov, Yaniv Markovski, Bianca Martin, Katie Mayer, Andrew Mayne, Bob McGrew, Scott~Mayer McKinney, Christine McLeavey, Paul McMillan, Jake McNeil, David Medina, Aalok Mehta, Jacob Menick, Luke Metz, Andrey Mishchenko, Pamela Mishkin, Vinnie Monaco, Evan Morikawa, Daniel Mossing, Tong Mu, Mira Murati, Oleg Murk, David Mély, Ashvin Nair, Reiichiro Nakano, Rajeev Nayak, Arvind Neelakantan, Richard Ngo, Hyeonwoo Noh, Long Ouyang, Cullen O'Keefe, Jakub Pachocki, Alex Paino, Joe Palermo, Ashley Pantuliano, Giambattista Parascandolo, Joel Parish, Emy Parparita, Alex Passos, Mikhail Pavlov, Andrew Peng, Adam Perelman, Filipe de~Avila Belbute~Peres, Michael Petrov, Henrique~Ponde de~Oliveira~Pinto, Michael, Pokorny, Michelle Pokrass, Vitchyr Pong, Tolly Powell, Alethea Power, Boris Power, Elizabeth Proehl, Raul Puri, Alec Radford, Jack Rae, Aditya Ramesh, Cameron Raymond, Francis Real, Kendra Rimbach, Carl Ross, Bob Rotsted, Henri Roussez, Nick Ryder,
  Mario Saltarelli, Ted Sanders, Shibani Santurkar, Girish Sastry, Heather Schmidt, David Schnurr, John Schulman, Daniel Selsam, Kyla Sheppard, Toki Sherbakov, Jessica Shieh, Sarah Shoker, Pranav Shyam, Szymon Sidor, Eric Sigler, Maddie Simens, Jordan Sitkin, Katarina Slama, Ian Sohl, Benjamin Sokolowsky, Yang Song, Natalie Staudacher, Felipe~Petroski Such, Natalie Summers, Ilya Sutskever, Jie Tang, Nikolas Tezak, Madeleine Thompson, Phil Tillet, Amin Tootoonchian, Elizabeth Tseng, Preston Tuggle, Nick Turley, Jerry Tworek, Juan Felipe~Cerón Uribe, Andrea Vallone, Arun Vijayvergiya, Chelsea Voss, Carroll Wainwright, Justin~Jay Wang, Alvin Wang, Ben Wang, Jonathan Ward, Jason Wei, CJ~Weinmann, Akila Welihinda, Peter Welinder, Jiayi Weng, Lilian Weng, Matt Wiethoff, Dave Willner, Clemens Winter, Samuel Wolrich, Hannah Wong, Lauren Workman, Sherwin Wu, Jeff Wu, Michael Wu, Kai Xiao, Tao Xu, Sarah Yoo, Kevin Yu, Qiming Yuan, Wojciech Zaremba, Rowan Zellers, Chong Zhang, Marvin Zhang, Shengjia Zhao, Tianhao
  Zheng, Juntang Zhuang, William Zhuk, and Barret Zoph. 2023.
\newblock \href {https://arxiv.org/abs/2303.08774} {Gpt-4 technical report}.
\newblock \emph{Preprint}, arXiv:2303.08774.

\bibitem[{Petroni et~al.(2019)Petroni, Rockt{\"a}schel, Riedel, Lewis, Bakhtin, Wu, and Miller}]{petroni-etal-2019-language}
Fabio Petroni, Tim Rockt{\"a}schel, Sebastian Riedel, Patrick Lewis, Anton Bakhtin, Yuxiang Wu, and Alexander Miller. 2019.
\newblock \href {https://doi.org/10.18653/v1/D19-1250} {Language models as knowledge bases?}
\newblock In \emph{Proceedings of the 2019 Conference on Empirical Methods in Natural Language Processing and the 9th International Joint Conference on Natural Language Processing (EMNLP-IJCNLP)}, pages 2463--2473, Hong Kong, China. Association for Computational Linguistics.

\bibitem[{Popovic(2017)}]{chrf++}
Maja Popovic. 2017.
\newblock \href {https://doi.org/10.18653/V1/W17-4770} {chrf++: words helping character n-grams}.
\newblock In \emph{Proceedings of the Second Conference on Machine Translation, {WMT} 2017, Copenhagen, Denmark, September 7-8, 2017}, pages 612--618. Association for Computational Linguistics.

\bibitem[{Qi et~al.(2023)Qi, Fern{\'a}ndez, and Bisazza}]{qi-etal-2023-cross}
Jirui Qi, Raquel Fern{\'a}ndez, and Arianna Bisazza. 2023.
\newblock \href {https://doi.org/10.18653/v1/2023.emnlp-main.658} {Cross-lingual consistency of factual knowledge in multilingual language models}.
\newblock In \emph{Proceedings of the 2023 Conference on Empirical Methods in Natural Language Processing}, pages 10650--10666, Singapore. Association for Computational Linguistics.

\bibitem[{Team(2022)}]{nllb2022}
NLLB Team. 2022.
\newblock No language left behind: Scaling human-centered machine translation.

\bibitem[{Touvron et~al.(2023{\natexlab{a}})Touvron, Martin, Stone, Albert, Almahairi, Babaei, Bashlykov, Batra, Bhargava, Bhosale, Bikel, Blecher, Canton{-}Ferrer, Chen, Cucurull, Esiobu, Fernandes, Fu, Fu, Fuller, Gao, Goswami, Goyal, Hartshorn, Hosseini, Hou, Inan, Kardas, Kerkez, Khabsa, Kloumann, Korenev, Koura, Lachaux, Lavril, Lee, Liskovich, Lu, Mao, Martinet, Mihaylov, Mishra, Molybog, Nie, Poulton, Reizenstein, Rungta, Saladi, Schelten, Silva, Smith, Subramanian, Tan, Tang, Taylor, Williams, Kuan, Xu, Yan, Zarov, Zhang, Fan, Kambadur, Narang, Rodriguez, Stojnic, Edunov, and Scialom}]{llama2}
Hugo Touvron, Louis Martin, Kevin Stone, Peter Albert, Amjad Almahairi, Yasmine Babaei, Nikolay Bashlykov, Soumya Batra, Prajjwal Bhargava, Shruti Bhosale, Dan Bikel, Lukas Blecher, Cristian Canton{-}Ferrer, Moya Chen, Guillem Cucurull, David Esiobu, Jude Fernandes, Jeremy Fu, Wenyin Fu, Brian Fuller, Cynthia Gao, Vedanuj Goswami, Naman Goyal, Anthony Hartshorn, Saghar Hosseini, Rui Hou, Hakan Inan, Marcin Kardas, Viktor Kerkez, Madian Khabsa, Isabel Kloumann, Artem Korenev, Punit~Singh Koura, Marie{-}Anne Lachaux, Thibaut Lavril, Jenya Lee, Diana Liskovich, Yinghai Lu, Yuning Mao, Xavier Martinet, Todor Mihaylov, Pushkar Mishra, Igor Molybog, Yixin Nie, Andrew Poulton, Jeremy Reizenstein, Rashi Rungta, Kalyan Saladi, Alan Schelten, Ruan Silva, Eric~Michael Smith, Ranjan Subramanian, Xiaoqing~Ellen Tan, Binh Tang, Ross Taylor, Adina Williams, Jian~Xiang Kuan, Puxin Xu, Zheng Yan, Iliyan Zarov, Yuchen Zhang, Angela Fan, Melanie Kambadur, Sharan Narang, Aur{\'{e}}lien Rodriguez, Robert Stojnic, Sergey Edunov,
  and Thomas Scialom. 2023{\natexlab{a}}.
\newblock \href {https://doi.org/10.48550/ARXIV.2307.09288} {Llama 2: Open foundation and fine-tuned chat models}.
\newblock \emph{CoRR}, abs/2307.09288.

\bibitem[{Touvron et~al.(2023{\natexlab{b}})Touvron, Martin, Stone, Albert, Almahairi, Babaei, Bashlykov, Batra, Bhargava, Bhosale et~al.}]{touvron2023llama}
Hugo Touvron, Louis Martin, Kevin Stone, Peter Albert, Amjad Almahairi, Yasmine Babaei, Nikolay Bashlykov, Soumya Batra, Prajjwal Bhargava, Shruti Bhosale, et~al. 2023{\natexlab{b}}.
\newblock Llama 2: Open foundation and fine-tuned chat models.
\newblock \emph{arXiv preprint arXiv:2307.09288}.

\bibitem[{Yang et~al.(2023)Yang, Xiao, Wang, Zhang, Bian, Yin, Lv, Pan, Wang, Yan et~al.}]{yang2023baichuan}
Aiyuan Yang, Bin Xiao, Bingning Wang, Borong Zhang, Ce~Bian, Chao Yin, Chenxu Lv, Da~Pan, Dian Wang, Dong Yan, et~al. 2023.
\newblock Baichuan 2: Open large-scale language models.
\newblock \emph{arXiv preprint arXiv:2309.10305}.

\end{thebibliography}



\end{CJK}
\end{document}